\documentclass[10pt]{article}
\pdfoutput=1

\usepackage{amsmath,amssymb,amsfonts}
\usepackage{algorithmic}
\usepackage{graphicx}
\usepackage{textcomp}
\usepackage{mathtools}
\usepackage{xcolor}
\usepackage{amsthm}
\usepackage{tabularx}
\usepackage{siunitx}
\usepackage{bm}
 \usepackage{multirow}
 \usepackage[toc]{appendix}
\usepackage[font=small]{caption}
\usepackage[caption=false]{subfig}

\usepackage{bm}
\def\BibTeX{{\rm B\kern-.05em{\sc i\kern-.025em b}\kern-.08em
    T\kern-.1667em\lower.7ex\hbox{E}\kern-.125emX}}
\usepackage{microtype}
\usepackage{booktabs}
\usepackage{hyperref}
\usepackage{natbib}
\usepackage[ruled,vlined,linesnumbered]{algorithm2e}

\newcommand{\predh}{{N_p}}
\newcommand{\ctrlh}{{N_c}}

\newcommand{\superviseMPC}{{\pi^{*}}}
\newcommand{\teacherMPC}{{\pi^{T}}}
\newcommand{\controlpolicy}{{\pi^L_\theta}}

\newcommand{\actionSuperviseMPC}{{u^{*}_t}}
\newcommand{\actionTeacherMPC}{{u^T_{t}}}
\newcommand{\actionControlpolicy}{{u^L_t}}

\newcommand{\disturbance}{d}

\usepackage[accepted]{icml2019}

\icmltitlerunning{}

\begin{document}

\twocolumn[
\icmltitle{MPC-guided Imitation Learning of Neural Network\\
Policies for the Artificial Pancreas}



\icmlsetsymbol{equal}{*}

\begin{icmlauthorlist}
\icmlauthor{Hongkai Chen}{sbece}
\icmlauthor{Nicola Paoletti}{rhul}
\icmlauthor{Scott A. Smolka}{sbcs}
\icmlauthor{Shan Lin}{sbece}
\end{icmlauthorlist}

\icmlaffiliation{sbece}{Department of Electrical and Computer Engineering, Stony Brook University, New York, USA}
\icmlaffiliation{rhul}{Department of Computer Science, University of London, Royal Holloway, UK}
\icmlaffiliation{sbcs}{Department of Computer Science, Stony Brook University, New York, USA}

\icmlcorrespondingauthor{Hongkai Chen}{hongkai.chen@stonybrook.edu}
\icmlcorrespondingauthor{Shan Lin}{shan.x.lin@stonybrook.edu}

\icmlkeywords{Machine Learning, ICML}

\vskip 0.3in
]



\printAffiliationsAndNotice{}  


\begin{abstract}

Even though model predictive control (MPC) is currently the main algorithm for insulin control in the artificial pancreas (AP), it usually requires complex online optimizations, which are infeasible for resource-constrained medical devices.  MPC also typically relies on state estimation, an error-prone process. In this paper, we introduce a novel approach to AP control that uses Imitation Learning to synthesize neural-network insulin policies from MPC-computed demonstrations. Such policies are computationally efficient and, by instrumenting MPC at training time with full state information, they can directly map measurements into optimal therapy decisions, thus bypassing state estimation. We apply Bayesian inference via Monte Carlo Dropout to learn
policies, which allows us to quantify prediction uncertainty and thereby derive safer therapy decisions.  We show that our control policies trained under a specific patient model readily generalize (in terms of model parameters and disturbance distributions) to patient cohorts, consistently outperforming traditional MPC with state estimation.

\end{abstract}


\section{Introduction}


The \textit{artificial pancreas} (AP) is a system for the automated delivery of insulin therapy for Type~1 diabetes~(T1D), a disease in which patients produce little or no insulin to regulate their blood glucose (BG) levels and maintain adequate glucose uptake in muscle and adipose tissue. The AP consists of an insulin infusion pump, and a subcutaneous Continuous Glucose Monitor (CGM) for sensing glucose levels.  CGM readings are transmitted to a control algorithm that computes the appropriate insulin dose. Such control should maintain a fine balance. Lack of insulin leads to hyperglycemia~(i.e., high BG), which, if untreated, can cause complications such as stroke, kidney failure, and blindness. Excessive insulin can lead to hypoglycemia (low BG), a critical state which can result in unconsciousness and death. 

Driven by advances in the mathematical modeling of T1D physiology~\cite{hovorka2004nonlinear,man2014uva}, \emph{Model Predictive Control} (MPC)~\cite{camacho2013model} has become the gold-standard AP algorithm.  MPC works by determining the insulin therapy that optimizes the future BG profile, predicted via physiological models. MPC, however, has two important limitations. First, it requires complex (often nonlinear and nonconvex) online optimization, which is infeasible when the algorithm is deployed in resource-constrained
medical devices. This is why commercial AP systems either use simplistic linear models within MPC\footnote{The algorithm of Typezero\circledR~\cite{forlenza2019successful} uses a linearized version of the nonlinear model of~\cite{man2014uva}.} or do not employ MPC at all in favor of simpler control algorithms (e.g., PID control in the Medtronic\texttrademark\ Minimed 670G~\cite{steil2013algorithms}, or fuzzy logic in the Glucositter by DreaMed Diabetes~\cite{atlas2010md}).

Second, and most crucial, MPC requires 
\emph{state estimation} (SE)
to recover the most recent patient state from CGM measurements~\cite{gondhalekar2014moving,paoletti2019data}. Besides its computational cost, SE is error-prone, as it relies strictly on CGM readings, which are an approximate, delayed, and noisy proxy of the target control variable, the BG. Incorrect state estimates might compromise the safety of the insulin therapy. 

\paragraph{Our Contributions.}
We present a novel method to derive \emph{end-to-end insulin control policies}, i.e., policies that subsume state estimation and control, directly mapping CGM measurements into patient-optimal insulin dosages. 
To capture the complex logic of MPC and SE, we consider policies represented as Deep Neural Networks, LSTMs in particular~\cite{hochreiter1997long}. Such an approach addresses the main issues surrounding the use of MPC, as it bypasses explicit SE and avoids the cost of MPC's online optimizations. 

Our approach is centered around the use of \emph{Imitation Learning} (IL)~\cite{ross2011reduction}, where the control policy is trained on examples provided by an MPC-guided teacher policy. Building on the PLATO framework~\cite{kahn2017plato}, at training time, we instrument the MPC teacher to access the full state of the patient model. As such, the learner policy, using only CGM measurements, will learn to mimic MPC-based decisions, which are based on the true model state and thus highly accurate. In this way, the learned policy incorporates both SE and control.

We employ IL as it alleviates the \emph{covariate shift}~\cite{storkey2009training} caused by the discrepancy between the training-time state distribution (induced by the teacher policy) and the test-time one (induced by the learner). Without IL, such a distribution shift can lead to unpredictable behaviour of the learner, and hence jeopardize the patient's safety.

We learn our insulin policies via approximate Bayesian inference, using \emph{Monte Carlo dropout}~\cite{gal2016dropout}. The resulting stochastic policy can capture model (epistemic) uncertainty, as well as uncertainty in the data, due to, e.g., noisy measurements~\cite{gal2016uncertainty}. Uncertainty quantification is key in safety-critical domains like ours, and is not addressed in the PLATO framework.  Variations in the patient's behavior and in physiological dynamics are indeed the norm, and can greatly affect the performance of a deterministic policy, with health consequences for the patient. In contrast, in a Bayesian framework, such variations likely yield high predictive uncertainty, information that we actively use to make robust therapy decisions. 

In summary, the main contribution of this paper is an Imitation Learning-based method for deriving Bayesian neural network policies for AP control, a method that overcomes the two main shortcomings of established MPC-based approaches, namely, SE errors and computational cost. 
We empirically show that: 1)~our IL-based approach outperforms Supervised Learning, and with less supervision data; 2)~the learned stochastic policies outperform MPC with SE and deterministic policies; 3)~our stochastic policies generalize to never-before-seen disturbance distributions and patient parameters that arise in a virtual patient cohort, while in the same setting, MPC with SE show consistent performance degradation. Overall, our best stochastic policy keeps BG in the safe range 8.4\%--11.75\% longer than MPC with SE and 2.94\%--9.07\% longer than the deterministic policy.

\section{Background on MPC for the AP}\label{sec:MPC_AP}

We consider an \textit{in silico} AP system, where the T1D patient is represented by a differential-equation model of BG and insulin metabolism, including absorption, excretion and transport dynamics between different body compartments. In particular, we chose the well-established model of~\citet{hovorka2004nonlinear}, a nonlinear ODE model with 14 state variables. 
A diagram of the MPC-based AP system is showed in Fig.~\ref{fig:system_test} (a). The state-space description is given in equations~(\ref{eq:state_space_form}--\ref{eq:mpc_pol}) below. The notation $a_{i,\ldots,i+j}$ stands for the indexed sequence $a_i,a_{i+1},\ldots ,a_{i+j}$.  

\begin{align}
\mathbf{x}_{t+1} = ~& {\bf F}_{\lambda_t}\left(\mathbf{x}_t, u_t, \disturbance_t \right), \  \lambda_t\sim \Lambda(\lambda \mid t), \disturbance_t\sim \mathcal{D}(d \mid t)
\label{eq:state_space_form}\\
y_t = ~& h\left(\mathbf{x}_t\right)+\epsilon_t \label{eq:output}\\
u_t = ~& \pi^*(\hat{\mathbf{x}}_t, \disturbance_{t,\ldots, t+N_p})\label{eq:mpc_pol}\\
\hat{\mathbf{x}}_t = ~& g(y_{t-N_b,\ldots,t}, u_{t-N_b,\ldots,t}, \disturbance_{t-N_b,\ldots,t-1}) \label{eq:state_estim}
\end{align}

Equation~\eqref{eq:state_space_form} describes the T1D patient model, where $\mathbf{x}_t$ is the patient state at time $t$, $u_t$ is the insulin input, $\lambda_t$ are the patient parameters with distribution $\Lambda(\lambda \mid t)$, and $\disturbance_t$ is the meal disturbance (amount of ingested carbohydrate) with distribution $\mathcal{D}(d \mid t)$. 
Note that both parameters and disturbances are random and time-dependent. Meal disturbances indeed depend on the patient's eating behaviour, which follows daily and weekly patterns. Similarly, parameters are typically subject to intra-patient variations such as daily oscillations. The parameter distribution $\Lambda(\lambda \mid t)$ can be alternatively defined to capture a population of patients (as we do in the experiment of Section~\ref{sec:evaluation}). 

Variable $y_t$ is the observed output at time $t$, i.e., the CGM measurement, which is subject to Gaussian sensor noise $\epsilon_t \sim \mathcal{N}(0,\sigma_{\epsilon})$, as done in previous work~\cite{soru2012mpc,patek2007linear}. With $\pi^*$ we denote the MPC-based control policy, that, given state estimate $\hat{\mathbf{x}}_t$ and future meal disturbances $\disturbance_{t,\ldots,t+N_p}$ as inputs, with $N_p$ being the MPC prediction horizon, computes the optimal insulin therapy by solving, at each time $t$, the following online optimization problem~\cite{paoletti2017data,paoletti2019data}\footnote{At each time step, the solution is actually a sequence of $N_p$ insulin values, of which we retain only the first one.}: 
\begin{multline}
\underset{u_{t,\ldots,t+\predh-1}}{\min} J(\hat{\mathbf{x}}_t, \disturbance_{t,\ldots,t+N_p}, u_{t,\ldots,t+\predh-1}) =  \\ 
\sum_{k=1}^\predh  d_{BG}(\tilde{\mathbf{x}}_{t+k}) + \beta\cdot \sum_{k=0}^{\ctrlh - 1} (\Delta u_{t+k})^2 \label{eq:rob_mpc_prob}
\end{multline}
\noindent subject to
\begin{align}
& u_{t+k} \in D_{u}, \quad k=0,\ldots,\ctrlh -1 \label{eq:rob_mpc2}\\
& u_{t+k} = \bar{u}, \quad k=\ctrlh,\ldots,\predh -1\label{eq:rob_mpc4}\\
& \tilde{\bf x}_t = \hat{\bf x}_t \label{eq:rob_mpc5}\\
& \tilde{\mathbf{x}}_{t+k} = 
{\bf F}_{\lambda_{t+k}}(\tilde{\bf x}_{t+k},u_{t+k},\disturbance_{t+k}), \ k=0,\ldots,\predh -1\label{eq:rob_mpc6}
\end{align}

\noindent where $N_c\leq N_p$ is the control horizon; $D_{u}$ is the set of admissible insulin values; \eqref{eq:rob_mpc4} states that $u$ is fixed to the basal insulin rate $\bar{u}$ outside the control horizon. \eqref{eq:rob_mpc5} and \eqref{eq:rob_mpc6} describe how the predicted state $\tilde{\bf x}$ evolves from estimated state $\hat{\bf x}_t$ at time $t$ following the plant model~\eqref{eq:state_space_form}. The objective function contains two terms, the former designed to penalize deviations between predicted and target BG (via function $d_{BG}$), the latter to avoid abrupt changes of the insulin infusion rate. The coefficient $\beta>0$ is a hyper-parameter.  See~\cite{paoletti2017data,paoletti2019data} for more details. 
Note that in order to predict future BG profiles, MPC requires some information of the future disturbances. In this work, we assume that the true future disturbance values are known. Alternative MPC formulations with weaker assumptions~\cite{hughes2011anticipating,paoletti2019data} could be considered alternatively.  

Finally, \eqref{eq:state_estim} describes state estimation (performed by function $g$), which follows the so-called \textit{Moving Horizon Estimation (MHE)} method~\cite{rawlings2013moving}. MHE is the dual of MPC in that MHE also uses model predictions but, this time, to estimate the most likely state given a sequence of $N_b$ {past} measurements, control inputs, and disturbances. The estimate is the state that minimizes the discrepancy between observed CGM outputs and corresponding model predictions. As such, \textit{the quality of the estimate directly depends on the accuracy of the predictive model and the quality of the measurements}. Details of the MHE optimization problem for the AP can be found in~\cite{chen2019committed}.
Alternative SE methods include Kalman filters and particle filters~\cite{rawlings2013moving}, but these are subject to the same kinds of estimation errors as MHE.

\section{Overview of the Method}\label{section:overview}

\begin{figure}
\centering
\subfloat[]{\includegraphics[width= 0.23\textwidth]{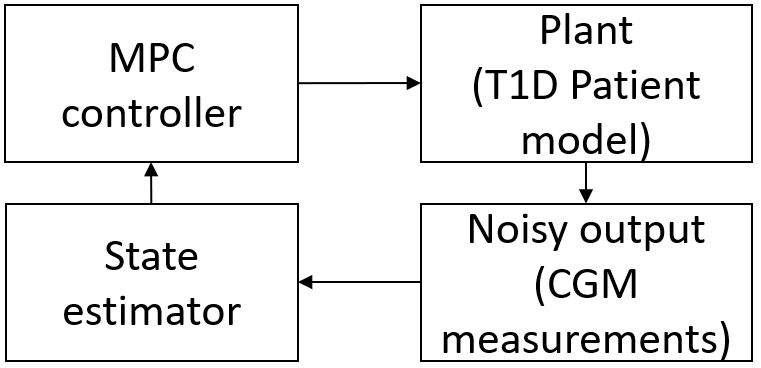}}
\hfill
\subfloat[]{\includegraphics[width= 0.23\textwidth]{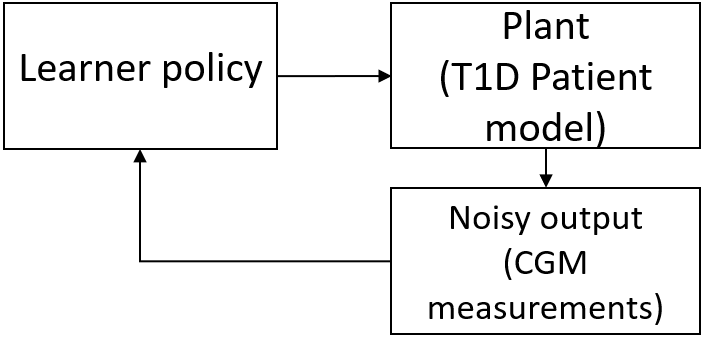}}
\vspace{-.3cm}
\caption{a)~A typical MPC-based AP system, where the controller requires a state estimate.  b)~Our learned end-to-end insulin policy instead only requires (noisy) observations.}
\vspace{-.3cm}
\label{fig:system_test}
\end{figure}

The main goal of our work is to design an end-to-end insulin control policy for the AP, i.e., policies that directly map noisy system outputs (CGM measurements) into an optimal insulin therapy, without requiring knowledge of the system state. The control loop of such a policy is illustrated in Fig.~\ref{fig:system_test} (b). To this purpose,  we follow an imitation learning approach where the \textit{learner policy} is trained from examples provided by an MPC-based expert, called the \textit{teacher policy}. 
Following the PLATO algorithm~\cite{kahn2017plato}, at training time, the teacher is instrumented with full state information, so that its demonstrations are not affected by SE errors.

\begin{figure}
\centering
\includegraphics[width=0.85\columnwidth]{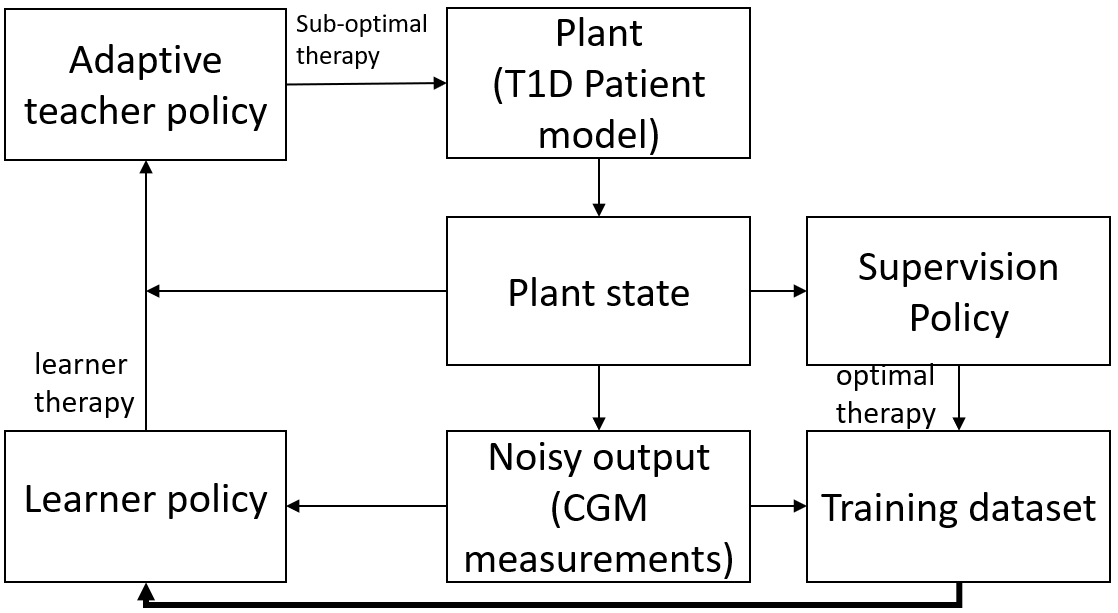}
\caption{Overview of the IL scheme for the AP. Training-time trajectories are generated by applying the adaptive teacher policy, which is a trade-off between the optimal supervision policy and the learner's policy. For training the learner, visited states are labelled with the optimal action of the supervision policy.}
\vspace{-.3cm}
\label{fig:systemOverview}
\end{figure}

Supervised learning (SL) is not a viable solution to our problem because SL assumes i.i.d.\ training and test data, while our case is subject to covariate shift: the training data follow the distribution of trajectories visited through the teacher, while at test time we have a different distribution, induced by the learner policy. If the learner cannot imitate perfectly the teacher, in the long-run the learner's actions will bring the system into a state far from the training distribution, where the behaviour of the learner becomes unpredictable, and with clear safety implications for our AP system. 
In IL, the teacher should provide demonstrations on trajectories that the learner would explore, but without knowing the learner in advance. To do so, a common solution is to reduce IL into a sequence of SL problems, where at each iteration, the learner is trained
on the distribution induced by the previous learners or by a ``mixture'' of learner and teacher policies~\cite{ross2010efficient,ross2011reduction}. 
 

Our IL method builds on the PLATO algorithm for adaptive trajectory optimization~\cite{kahn2017plato}. See Section~\ref{sec:relatedwork} for a summary of our extensions to PLATO. 
PLATO also reduces IL into a sequence of SL problems, where at each iteration, the teacher's actions gradually adapt to those of the current learner policy. 
This \textit{adaptive teacher} is an MPC-based policy whose objective function is extended to match the learner's behaviour. As such, it is non-optimal, and is only used to generate trajectories that approach the distribution induced by the learner, thereby alleviating the covariate shift problem. The training data is obtained by labelling the adaptive teacher trajectories with the original (optimal) MPC policy. 
We call the latter \textit{supervision policy}. The training process is summarized in Fig.~\ref{fig:systemOverview}. 
In our approach the learner policy is stochastic, represented as a Bayesian neural network, as explained in Section~\ref{sec:bayesian}. 

\section{Imitation Learning Algorithm}\label{section:algorithm}

The supervision policy $\superviseMPC$ is the MPC policy for the AP system of (\ref{eq:rob_mpc_prob}--\ref{eq:rob_mpc6}). Its control action at time $t$, $u^*_t$, is obtained by running the MPC algorithm given the true system state ${\mathbf{x}}_t$ (instead of the estimated state) and future meal disturbances: $u^*_t = \superviseMPC({\mathbf{x}}_t, \disturbance_{t\ldots t+N_p})$. 
The learner policy $\controlpolicy$ is represented by an LSTM network with parameters $\theta$. 
The choice of a recurrent architecture is natural for our application, because our policies have to subsume both control and SE and, as discussed in Section~\ref{sec:MPC_AP}, SE for nonlinear models can be seen as a sequence prediction problem, see~\eqref{eq:state_estim}. 

At time $t$, the learner policy takes 
an input sequence, which includes past observations $y_{1,\ldots,t}$,  past and future disturbances $d_{N_p+1,\ldots, t+N_p}$, and past control actions $u_{1,\ldots,t-1}$, and predicts the control action at time $t$, $u^L_t$, by iterating the following equation:
\vspace{-.1cm}
\begin{align}
    {\bf s}_{t} = & \ f_{\theta}({\bf s}_{t-1},u_{t-1},y_t,d_{t+N_p})\\
    \actionControlpolicy = & \ g_{\theta}({\bf s}_{t})\\
    \controlpolicy= & \ g_{\theta} \circ f_{\theta}
\end{align}
where ${\bf s}_{t}$ is the hidden state of the network at time $t$, and $f_{\theta}$ and $g_{\theta}$ are activation functions. The learner results from the composition of $g_{\theta}$ and $f_{\theta}$. Since we use a stochastic policy for $\controlpolicy$ (see Section~\ref{sec:bayesian}), we will denote the learner by the conditional distribution $\controlpolicy(\actionControlpolicy \mid {\bf s}_{t-1},u_{t-1},y_t,d_{t+N_p})$.

The adaptive teacher policy $\teacherMPC$ extends the supervision policy $\superviseMPC$ with a term to penalize the mismatch between the supervision policy, i.e., the optimal MPC policy, and the learner. The output $\actionTeacherMPC$ of $\teacherMPC$ is the first control action in the solution of the following MPC problem.
\begin{align}
\label{eq:teacher_prob}
& \underset{u^T_{t,\ldots,{t+\predh-1}}}{\min} J(\mathbf{x}_t, \disturbance_{t,\ldots ,t+N_p}, u^T_{t,\ldots,t+\predh-1}) + \rho \cdot J_M & \\[-.5cm]
& \text{subject to (\ref{eq:rob_mpc2}--\ref{eq:rob_mpc6}). } &\nonumber
\vspace*{-.3cm}
\end{align}

The first term of~\eqref{eq:teacher_prob} is the cost function of the supervision policy~\eqref{eq:rob_mpc_prob}. The second term penalizes the distance between the optimal (supervision) control action and the action that the learner policy would perform.  
The factor $\rho$ determines the relative importance of matching the behavior of the learner policy $\controlpolicy$ against minimizing the cost $J$. In our experiments, we set $\rho = 1- 0.8^{i-1}$ where $i$ is the iteration of our IL algorithm, see Algorithm~\ref{alg:PLATO}. In this way, the relative importance of matching $\controlpolicy$ increases as the learner improves (i.e., as $i$ increases). By gradually matching the behaviour of the learner, the control actions taken by $\teacherMPC$ will lead to exploring a state space similar to the one induced by the learner policy.

The distance term $J_M$ is defined as the $p$-th Wasserstein distance $W_p$ between the output distribution of the learner policy $\controlpolicy(\actionControlpolicy \mid {\bf s}_{t-1},u_{t-1},y_t,d_{t+N_p})$ 
and the optimal control action $u^T_t$, or more precisely, the Dirac distribution centered at $u_t^T$, $\delta_{u_t^T}$. We choose $W_p$ because the two distributions
are likely to have disjoint supports, and so other measures like the KL divergence (used in PLATO) are undefined or infinite. 
As we will see, an analytical form for $\controlpolicy$ is not available, but we can sample from it and obtain the empirical approximation $\frac{1}{n}\sum_{i=1}^n \delta_{u^L_{t,i}}$, where $u^L_{t,1},\cdots,u^L_{t,n}$ are the sampled values of $\controlpolicy$. Thus, we have that 
\begin{equation}\label{eq:wass}
            J_M =  W_p\left(\delta_{\actionTeacherMPC},\frac{1}{n}\sum_{i=1}^n \delta_{u^L_{t,i}}\right)\\
             = \left( \frac{1}{n} \sum_{i=1}^n d(u^L_{t,i},\actionTeacherMPC)^p \right)^{1/p},
    \vspace{-0.1cm}
\end{equation}
where $d$ is a suitable distance. Note that the above equality holds as we are comparing a discrete distribution with a Dirac distribution. We set $p=1$ and $d(\cdot)=\|\cdot \|_2$. 


Algorithm~\ref{alg:PLATO} outlines our IL scheme, which consists of a sequence of $N$ SL iterations. The learner policy is initialized at random. At each iteration, we start from a random initial state of the plant and generate a trajectory of the system of length $T_e$. To do so, we first sample a sequence of random meal disturbances, i.e., of ingested carbohydrates (lines 5-7).
Then, for each time point $t$ of the trajectory, the adaptive teacher $\teacherMPC$ computes an insulin infusion rate {$u^T_t$} by solving~\eqref{eq:teacher_prob}, that is, by optimizing the MPC objective while matching the current learner policy {${\pi^L_{\theta_i}}$} (line 11).
The MPC supervision policy $\superviseMPC$ is used to compute the optimal control action {$\actionSuperviseMPC$} by solving~\eqref{eq:rob_mpc_prob} (line 12).
The optimal action is used to label the corresponding training example $\left((y_t,u^T_{t-1},d_{t+\predh}),\actionSuperviseMPC\right)$, which is added to the training set $S$ (line 13), 
while the sub-optimal action by the adaptive teacher is used to evolve the system state (line 14). 
At the end of each iteration, {${\pi^L_{\theta_i}}$} is trained using the current set of examples $S$. As the teacher gradually matches the learner, the training-time distribution of the system state gradually approximates that at test time. 

The time complexity of Algorithm~\ref{alg:PLATO} is dominated by the two MPC instances (lines 11 and 12), which are solved $N\cdot T_e$ times, and the training of the learner $\pi^L$, repeated $N$ times.

\begin{algorithm}[tb]
   \caption{IL for Artificial Pancreas control policy using adaptive trajectory optimization}
   \label{alg:PLATO}
\begin{algorithmic}[1]
    
   \STATE Initialize training dataset $S\leftarrow \emptyset$.
   \STATE Randomly initialize learner policy $\pi^L_{\theta_1}$.
   \STATE Sample random patient parameters $\lambda \sim \Lambda(\lambda \mid t)$
   \FOR{$i=1$ {\bfseries to} $N$}
   \FOR{$t=1$ {\bfseries to} $T_e+N_p$}\label{lst:line:sample_disturbance1}
   \STATE Sample random carb disturbance $\disturbance_t \sim \mathcal{D}(d \mid t)$\label{lst:line:sample_disturbance2}
   \ENDFOR\label{lst:line:sample_disturbance3}
   \STATE Initialize the patient state ${\bf x}_1$.
    \FOR{$t=1$ {\bfseries to} $T_e$}
    \STATE Collect CGM measurements $y_t$ as per~\eqref{eq:output}
    \STATE  Compute sub-optimal therapy with adaptive teacher:  $\actionTeacherMPC \sim \teacherMPC(\actionTeacherMPC \mid {\bf x}_t,d_{t,\cdots,t+\predh}, {\pi^L_{\theta_i}})$.\label{lst:line:adaptive_action}
    \STATE  Compute optimal therapy with supervision policy: $\actionSuperviseMPC = \superviseMPC({\bf x}_t, d_{t,\cdots,t+\predh})$.\label{lst:line:supervised_action}
    \STATE Append $\left((y_t,u^T_{t-1},d_{t+\predh}),\actionSuperviseMPC\right)$ to $S$.\label{lst:line:append_data}
    \STATE State evolves as ${\bf x}_{t+1} =  {\bf F}_{\lambda}\left(\mathbf{x}_t, \actionTeacherMPC, \disturbance_t \right)$\label{lst:line:evolve_by_teacher} 
    \ENDFOR
    \STATE Train $\pi^L_{\theta_{i+1}}$ on $S$\label{lst:line:train_by_supervision}
   \ENDFOR

\end{algorithmic}
\end{algorithm}

\section{Bayesian inference of control policy}\label{sec:bayesian}
We take a Bayesian approach to learn our control policy, which results in a stochastic policy represented by a neural network with fixed architecture and random parameters $\theta$. This provides us with a distribution of policy actions, the \textit{predictive distribution}, from which we can derive uncertainty measures to inform the final therapy decision. Such uncertainty should capture both data uncertainty, due to e.g., noisy measurements, and epistemic uncertainty, i.e., the lack of confidence of the model about a given input~\cite{gal2016uncertainty}. 



Performing Bayesian inference corresponds, given training data $S$, to computing the posterior $p(\theta \mid S)$ from some prior $p(\theta)$ by applying Bayes rule. 
The distribution of policies is induced by the random parameters $\theta \sim p(\theta \mid S)$.  In particular, the predictive distribution 
$p(\actionControlpolicy \mid x, S)$, with $x=({\bf s}_{t-1},u_{t-1},y_t,d_{t+N_p})$ being the policy inputs, is obtained from $p(\theta \mid S)$ as:
\begin{equation}\label{eq:pred_dist}
p(\actionControlpolicy \mid x, S) = \int \controlpolicy(\actionControlpolicy \mid x) \cdot p(\theta \mid S) \ {\rm d}\theta.
\vspace*{-.3cm}
\end{equation}

For the non-linearity of the neural network function, precise inference is, however, infeasible and thus one needs to resort to approximate methods~\cite{neal2011mcmc,blundell2015weight}. 
\emph{Monte Carlo Dropout} (MCD)~\cite{gal2016dropout} is one of these. 
Dropout is a well-established regularization technique 
based on dropping some neurons at random during training with some probability $p$. Technically, this corresponds to
multiplying the weights with a \textit{dropout mask}, i.e., a vector of Bernoulli variables with parameter $p$. 
Then, at test time, standard dropout derives a deterministic network by scaling back the weights by a factor of $1/(1-p)$. 
On the other hand, in MCD the random dropout mask is applied at test time too, resulting in a distribution of network parameters. \citet{gal2016dropout} show that applying dropout to each weight layer is equivalent to performing approximate Bayesian inference of the neural network. This property is very appealing as it reduces the problem of inferring $p(\theta \mid S)$ to drawing Bernoulli samples, which is very efficient.

\paragraph{Sample size.} For the empirical approximation of~\eqref{eq:pred_dist}, we choose a sample size of $n=47$ that, by the Dvoretzky-Kiefer-Wolfowitz inequality~\cite{massart1990tight}, gives us the following statistical guarantee:
\begin{equation*}
    {\rm Pr}\left(\underset{x\in \mathbb{R}}{\sup} \ |F_n(x) - F(x)| > 0.2 \right) < 0.05,
\end{equation*}
where $F$ is the CDF of the predictive distribution and $F_n$ its empirical approximation with $n=47$ observations. 

\paragraph{Decision rule.} The output of our policy is the distribution of insulin actions~\eqref{eq:pred_dist}. Hence, we need to define a decision rule that produces an individual value $u^t$ out of this distribution, and we want such rule to account for the predictive uncertainty of the policy. Let $u^L_{t,1},\cdots,u^L_{t,n}$ be the ordered empirical sample of~\eqref{eq:pred_dist}, and let $y_{t-1}$ be the last measured glucose value. Our rule selects a particular order statistic $u^L_{t,M}$, i.e., one of the sampled values, depending on the relative distance of $y_{t-1}$ w.r.t.\ the safe BG upper bound $BG_{ub}$ and lower bound $BG_{lb}$. We call this \textit{adaptive rule} because the selected order statistic is adapted on $y_{t-1}$. Formally,
\begin{equation}\label{eq:adaptive}
u^t=u^L_{t,M}, \text{with } M=n \cdot \lceil (y_{t-1}-BG_{lb})/(BG_{ub} - BG_{lb})\rceil.
\end{equation}
In this way, if the patient is approaching hypoglycemia ($y_{t-1}$ close to $BG_{lb}$), we select a conservative insulin value, and we select instead an aggressive therapy if $y_{t-1}$ is close to hyperglycemia ($BG_{ub}$). Importantly, as the policy uncertainty increases (and so does the spread of the sample), $u^t$ gets more conservative when $y_{t-1}$ is in the lower half of the safe BG range, i.e., we take safer decisions when the policy is less trustworthy\footnote{Protecting against hypoglycemia is the primary  concern.}. For the same principle, when $y_{t-1}$ is in the upper half of the range, higher uncertainty yields a more aggressive therapy, but this poses no safety threats because $y_{t-1}$ is well away from the hypoglycemia threshold $BG_{lb}$. 

In our evaluation, we compare our adaptive rule with the commonly used rule that sets $u^t$ to the sample mean of~\eqref{eq:pred_dist}.

\section{Experiments}\label{sec:evaluation}
We conducted computational experiments to validate the following claims:
\begin{enumerate}
    \item Our IL-based approach outperforms SL, and with less supervision data.
    \item The learned stochastic policies outperform both MPC with SE and deterministic policies.
    \item The stochastic policies generalize well to unseen patient physiological parameters and meal disturbances.
    \item For a stochastic policy, the adaptive decision rule~\eqref{eq:adaptive} outperforms the mean prediction rule.
    \item The predictive uncertainty of the policy increases with out-of-distribution test inputs.
\end{enumerate}





\paragraph{Runtime cost.} On our workstation\footnote{With an Intel i7-8750H CPU and 16GB DDR4 SDRAM} the stochastic policy executes in approximately 20 milliseconds, which is well within the CGM measurement period of 5 minutes, and consistently more efficient than MPC-based optimization (on average, 150+ times faster in our experiments)\footnote{Optimization for MPC and SE is solved used MATLAB's implementation of the interior-point algorithm of~\citet{byrd2000trust}.}. 
Thanks to platforms such as TensorFlow Lite, we believe that a similar runtime can be obtained also after deploying the policy on embedded hardware. 


\paragraph{LSTM architecture.} 
We represent the learner policy as an LSTM network with three layers, of 200 hidden units each, with \texttt{tanh} state activation functions and \texttt{sigmod} gate activation functions. The network has a fully connected layer, a ReLu activation function layer and a regression layer as final layers. We added a dropout layer before each weight layer with dropout probability $0.2$, which empirically gave us the best accuracy among other probability values. 
We also experimented with feedforward architectures but these performed poorly, confirming the need for a recurrent architecture to adequately represent SE and control.


\paragraph{Performance measures.} To evaluate and compare the insulin policies, we consider three typical performance measures in the AP domain: $t_{hypo}, t_{hyper},t_{eu}$: the average percentage of time the virtual patient's BG is in hypoglycemia (i.e., BG $\leq$ 70 mg/dL), hyperglycemia (i.e., BG $\geq$ 180 mg/dL), and in the safe range, or euglycemia (i.e., 70 mg/dL $\leq$ BG $\leq$ 180 mg/dL), respectively. Given the goal of diabetes treatment and BG control, the target is to maximize $t_{eu}$ while minimizing $t_{hypo}$ (hypoglycemia leads to more serious consequences than hyperglycemia).







\paragraph{Experimental settings.}
During training, we consider a fixed (and time-constant) parameterization of the T1D virtual patient model, and the disturbance distribution of Table~\ref{table:meals1} (which includes three large meals and three small snacks). The length of each training trajectory is set to $T_e = 1440$. We performed $N = 34$ iterations of the IL algorithm, totalling to a training set size of $|S|=48,960$. 
At test time, we consider the following three configurations of the model parameters:
\begin{enumerate}
    \item \textit{Fixed Patient.} The test-time virtual patient has the same parameters as during training.
    \item \textit{Varying Patient.} We introduce intra-patient temporal variations to the parameters. 
    \item \textit{Patient cohort.} We introduce both inter-patient and intra-patient variations. The distribution of model parameters represents a cohort of virtual patients, where each patient has time-varying parameters. It might include severe cases when the  parameters are sampled at marginal areas of the distribution.
\end{enumerate}
The above distributions for modelling intra- and inter-patient variability are taken from~\cite{wilinska2010simulation}, where the authors derived them from clinical data. 

For each configuration and policy, we simulate 90 trajectories of a one-day simulation. For each trajectory, we draw a fresh observation of the random disturbance and patient parameters, but, to ensure a fair comparison, these are kept the same for all control algorithms. 

In another set of experiments, we consider (in combination with the above configurations) a meal disturbance distribution different from the one at training time. This meal distribution reflects a late eating habit and a larger probability of larger snacks, which leads to meal profiles with higher carbs. Further details are reported in the supplement. 

For each experiment, we evaluate the performance of the following policies:
\begin{itemize}
    \item \textbf{MPC+SI:} the supervision policy $\pi^*$, i.e., the MPC algorithm with full  information on the current state and model parameters. We stress that full observability is an unrealistic assumption, yet useful to establish an ideal performance level. {We set control and prediction horizons of MPC to $N_c = 100$ min and $N_p = 150$ min, respectively.}
    \item \textbf{MPC+SE:} the MPC algorithm with state estimation, as described in~(\ref{eq:rob_mpc_prob}--\ref{eq:rob_mpc6}). This is one of the most common control schemes for the AP. The parameters of the T1D model used within MPC and SE are set to the training-time values. We will show that SE takes a toll on the overall performance, unlike our end-to-end policies.
    \item \textbf{DLP:} the deterministic learner policy, where dropout is used at training time but not at test time. 
    \item \textbf{SLP-M:} the stochastic learner policy with the rule that selects the sample mean of the predictive distribution.
    \item \textbf{SLP-A:} the stochastic learner policy with the adaptive rule of~\eqref{eq:adaptive}. As explained above, such rule actively uses uncertainty information by selecting an order statistic of the predictive distribution commensurate to the sensed glucose. We will show that this policy outperforms all the others (but  {MPC+SI}). 
\end{itemize}

\paragraph{IL outperforms SL.} We compared the performance of {SLP-A}, which is trained via the IL algorithm of Section~\ref{section:algorithm}, against the corresponding stochastic policy trained using an MPC-guided SL approach, that is, without the adaptive teacher policy that guides the exploration of training trajectories but only using the supervision policy. 

\begin{figure*}
\centering
\subfloat[Fixed patient]{\includegraphics[width=.32\textwidth]{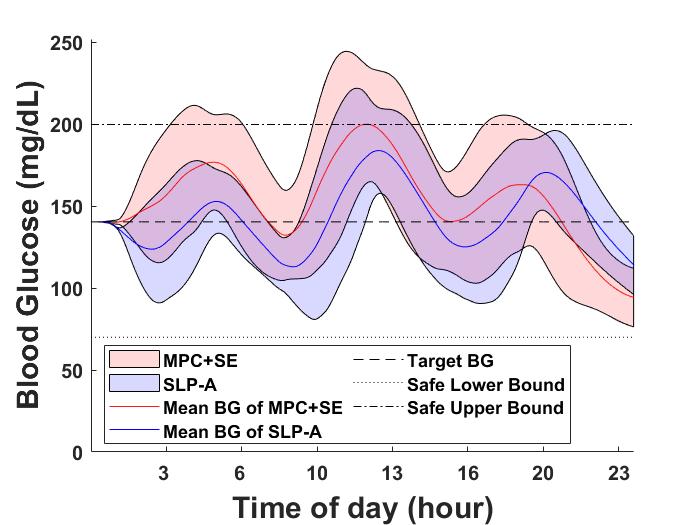}} \hfill
\subfloat[Varying patient]{\includegraphics[width=.32\textwidth]{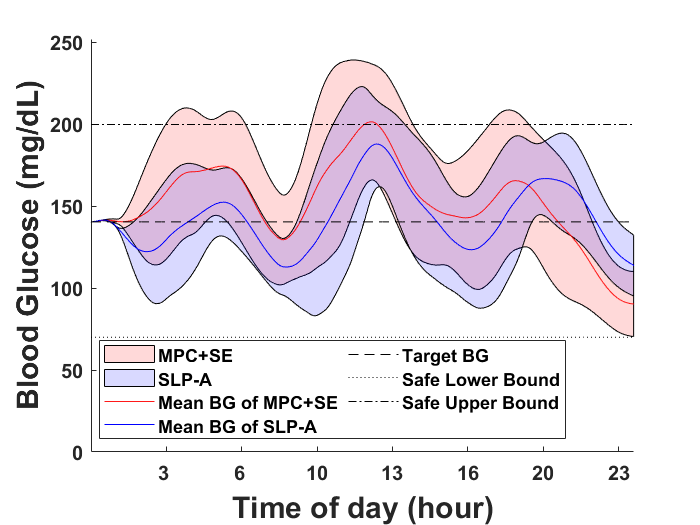}} \hfill
\subfloat[Patient cohort]{\includegraphics[width=.32\textwidth]{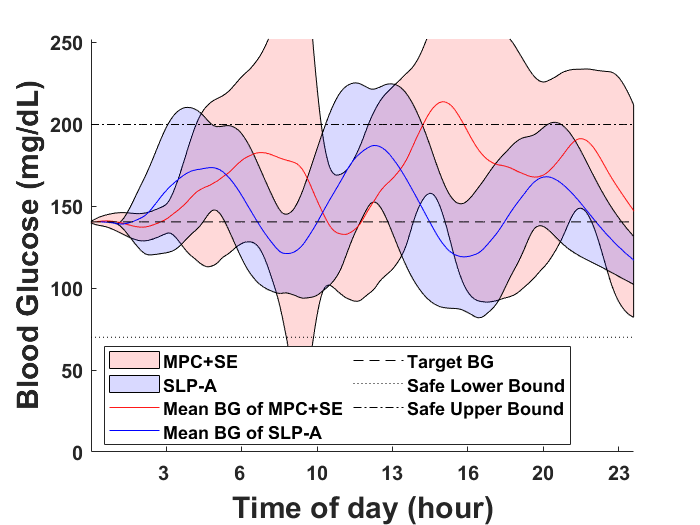}}
\caption{Mean BG $\pm$ standard deviation for MPC with state estimation vs stochastic policy with adaptive output.}
\vspace*{-.3cm}
\label{fig:bg_profiles}
\end{figure*}

\begin{table}[htbp]
\large
\centering
\caption{Attributes of meal disturbance distribution during training. CHO amounts and starting times are sampled uniformly from the reported intervals.}
\vskip 0.1in
\label{table:meals1}
\resizebox{0.5\textwidth}{!}{%
\begin{tabular}{lllllll}
\hline
                 & breakfast              & snack 1               & lunch                   & snack 2               & dinner                 & snack 3               \\ \hline
Occurrence       & \multirow{2}{*}{100}   & \multirow{2}{*}{50}   & \multirow{2}{*}{100}    & \multirow{2}{*}{50}   & \multirow{2}{*}{100}   & \multirow{2}{*}{50}   \\
Probability (\%) &                        &                       &                         &                       &                        &                       \\ \hline
CHO Amount       & \multirow{2}{*}{40-60} & \multirow{2}{*}{5-25} & \multirow{2}{*}{70-110} & \multirow{2}{*}{5-25} & \multirow{2}{*}{55-75} & \multirow{2}{*}{5-15} \\
(gram)           &                        &                       &                         &                       &                        &                       \\ \hline
Time of          & 1:00-                  & 5:00-                 & 8:00-                   & 12:00-                & 15:00-                 & 19:00-                \\
day (hour)       & 5:00                   & 8:00                  & 12:00                   & 15:00                 & 19:00                  & 21:00                 \\ \hline
\end{tabular}%
}
\end{table}

We found that the IL-based policy has much better control performance than the SL-based one, using the same number of algorithm iterations and training examples. This suggests that the IL-based policy captures more useful state-action information and that the IL algorithm leads the learner policy to efficiently explore the state space. In particular, after 20 iterations of Algorithm~\ref{alg:PLATO}, {SLP-A} achieves an average of $79.60\%$ for $t_{eu}$ (time in range), while the SL-based counterpart only $17.35\%$. IL consistently outperforms SL in all other performance metrics too (see Appendix~\ref{app:SL} for more detailed results).

\paragraph{Stochastic policies outperform MPC with SE and deterministic policies.} Results in Table~\ref{tab:my-table} evidence that the stochastic policies ({SLP-A} and {SLP-M}) outperform the MPC policy with state estimation in essentially all performance measures and configurations, and in a statistically significant manner. In particular, we observe that, on average, the {SLP-A} policy stays in range for 8.4\%--11.75\% longer than {MPC+SE}, and the {SLP-M} does so for 5.97\%--8.4\% longer. 
This is visible also from the BG profiles of Figure~\ref{fig:bg_profiles}. 
The deterministic policy also performs better than {MPC+SE}, but lags behind the stochastic ones, with {SLP-A} achieving a time in range 2.94\%--9.07\% longer than {DLP}. 

These results suggests that MPC with SE introduces estimation errors that have a detrimental effect on BG control, as also confirmed by the fact that the same control algorithm but with full state information ({MPC+SI}) is way superior. In realistic settings where the true patient state is not accessible, our analysis shows that an end-to-end policy is to be preferred to explicit SE. 


\vspace*{-.3cm}
\paragraph{Stochastic policies generalize to unseen patient parameters and disturbances.}
From Table~\ref{tab:my-table} and Figure~\ref{fig:bg_profiles}, we observe that our stochastic policies are robust to patient parameters unseen during training, with a time in range constantly above 85\% for the varying patient and the patient cohort configurations. 
The superiority of the stochastic policies over the deterministic one under intra-patient and inter-patient variability evidence the role of considering prediction uncertainty when the policy is deployed in environments that deviate from the training one.

\begin{table*}[tbp]
\centering
\caption{Performance of MPC with state information (MPC+SI), MPC with state estimation (MPC+SE), deterministic learner policy (DLP), stochastic learner policy with mean value output (SLP-M), and stochastic learner policy with adaptive output (SLP-A). For each column, in bold are the significantly best policies, i.e., with $p<0.005$ in all pairwise sign tests (one-sided)~\cite{hollander2013nonparametric}. A complete, detailed table of performance statistics and the processes of hypothesis tests can be found in Appendix~\ref{app:sgtest}.}
\vskip 0.1in
\label{tab:my-table}
\resizebox{\linewidth}{!}{%
\begin{tabular}{@{}llllllllll@{}}
\toprule
       & \multicolumn{3}{l}{Fixed Patient Configuration}           & \multicolumn{3}{l}{Varying Patient Configuration}         & \multicolumn{3}{l}{Patient Cohort Configuration}          \\ \midrule
       & $t_{hypo}$ (\%) & $t_{eu}$ (\%) & $t_{hyper}$ (\%) & $t_{hypo}$ (\%) & $t_{eu}$ (\%) & $t_{hyper}$ (\%) & $t_{hypo}$ (\%) & $t_{eu}$ (\%) & $t_{hyper}$ (\%) \\ \midrule
MPC+SI & 0.00  $\pm$ 0.00      & 99.84$\pm$ 0.60      & 0.16$\pm$ 0.60         & 0.00$\pm$0.00         & 99.80$\pm$0.83       & 0.20$\pm$0.83          & 0.00$\pm$0.00         & 99.15$\pm$1.81       & 0.85$\pm$1.81          \\ \midrule
MPC+SE & 0.92$\pm$2.76         & 80.40$\pm$5.33       & 18.68$\pm$4.76         & 1.44$\pm$3.18         & 79.88$\pm$5.46       & 18.69$\pm$4.95         & 0.85$\pm$3.57         & 77.17$\pm$15.39      & 21.98$\pm$14.30        \\ \midrule
DLP    & 0.21$\pm$0.99         & 85.40$\pm$4.15       & 14.38$\pm$3.90         & 0.53$\pm$1.87         & 82.66$\pm$6.42       & 16.82$\pm$6.25         & 0.45$\pm$1.91         & 82.63$\pm$6.28       & 16.93$\pm$6.15         \\ \midrule
SLP-M    & 0.23$\pm$1.10         & {86.33$\pm$4.17}       & {13.44$\pm$4.09}         & 0.33$\pm$1.09         & {86.34$\pm$4.20}       & {13.33$\pm$4.06}         & 0.32$\pm$1.19         & \textbf{85.45$\pm$4.34}       & \textbf{14.23$\pm$4.00}         \\ \midrule
SLP-A    & 0.13$\pm$0.64         & \textbf{91.73$\pm$3.45}       & \textbf{8.14$\pm$3.39}          & 0.05$\pm$0.27         & \textbf{91.73$\pm$3.18}       & \textbf{8.23$\pm$3.15}          & 0.04$\pm$0.41         & \textbf{85.57$\pm$4.12}       & \textbf{14.39$\pm$4.08}         \\ \bottomrule
\end{tabular}%
}
\end{table*}

We further evaluate the robustness of the SLP-A policy under an unseen meal disturbance distribution characterized by a higher carbs intake. Results, reported in Table~\ref{table:meals1_result}, evidence that our stochastic policy generalize well also in this case. A hypothesis test can be found in Appendix~\ref{app:meal}, showing that the performance statistics  SLP-A achieves under intra-patient and inter-patient variability with two different meal distributions has no significant difference.

\paragraph{Adaptive rule outperforms mean-value rule.} The SLP-A policy obtains a time in range approximately 5\% longer than SLP-M in the fixed patient and varying patient configurations, see Table~\ref{tab:my-table}. There is an improvement also in the patient cohort experiment, albeit less significant. With the adaptive rule, the policy adopts a more conservative or aggressive therapy depending on the measured glucose and the predictive uncertainty, which can lead to more stable BG trajectories and explain the observed difference.

\paragraph{Policy uncertainty increases at out-of-distribution test inputs.} We verify that inference via Monte Carlo dropout adequately captures epistemic uncertainty. We indeed observe an increased variability of the predictive distribution in the patient cohort configuration, that is, when the policy is applied to out-of-distribution test data resulting from introducing inter- and intra-patient variability. However, uncertainty does not significantly increase from the first to the second configuration, that is, when introducing only inter-patient variability. These results were confirmed via pair-wise two-sample Kolmogorov-Smirnov tests over the coefficients of variation (CoVs),  i.e., the mean-normalized standard deviations (differences significant at $p<10^{-37}$). See Appendix~\ref{app:uncertain} for the elaboration on the Kolmogorov-Smirnov tests and a plot of the CoVs distributions.

\begin{table}[tbp]
\small
\centering
\caption{Performance of SLP-A with unseen disturbances.}
\vskip 0.1in
\label{table:meals1_result}
\resizebox{0.5\textwidth}{!}{%
\begin{tabular}{@{}lllllll@{}}
\toprule
{Performance }&{Fixed Patient} &{Varying Patient} &{Patient Cohort} \\
Metrics &  Configuration &  Configuration &  Configuration\\ \midrule
{$t_{hypo}$ ($\%$)}& {$0.35\pm 1.08$ }&{$0.67\pm 1.83$}&{$0.00\pm 0.00$}\\\midrule

{$t_{eu}$ ($\%$)}&{$91.41 \pm 3.67$}&{$90.77\pm 4.44$}&{$85.92\pm 3.96$}\\\midrule

{$t_{hyper}$ ($\%$)}&{$8.24 \pm 3.43$}&{$8.56\pm 3.73$}&{$14.08 \pm 3.96$}\\\bottomrule
\end{tabular}%
}
\vspace{-0.2in}
\end{table}

\section{Related Work} \label{sec:relatedwork}

Traditional methods for insulin control in the AP mostly rely on  MPC~\cite{hovorka2004nonlinear,magni2007model,lee2009closed,cameron2011closed,paoletti2019data}, PID control~\cite{steil2004closed,huyett2015design}, and fuzzy rules based on the reasoning of diabetes caregivers~\cite{atlas2010md,nimri2012feasibility}. 
Reinforcement learning approaches have been proposed as well, including policy iteration~\cite{de2015controlling,weng2017representation}, actor critic methods~\cite{articledsfa}, and deep Q-networks for dual-hormone therapy~\cite{zhu2019dual}\footnote{Dual-hormone therapy allows infusion of both insulin and its antagonist hormone, glucagon, which protects against hypoglycemia. However, it is not as established as insulin-only therapy, and still has safety concerns~\cite{taleb2017glucagon}.}. 
Neural networks have been explored in the AP space not just to represent insulin policies~\cite{de2012artificial,zhu2019dual}, but also to predict BG concentration based on inputs such as insulin dosage, nutritional intake, and daily activities~\cite{pappada2008development, perez2010artificial,dutta2018robust,li2019glunet}. Our work is different from the above papers as: 1) it takes an imitation learning approach to learn the insulin policy, mitigating potential (and dangerous) test-time distribution drifts; 2) incorporates in the policy uncertainty information obtained via Bayesian inference; 3) it produces end-to-end policies that do not require learning a separate BG prediction model.

Several imitation learning methods have been proposed, such as~\cite{daume2009search,ross2010efficient,articlwefae,ho2016generative}. Some of these, like our approach, are tailored to work with MPC teachers~\cite{inproceedinghjjgs,kahn2017plato,Amos:2018:DME:3327757.3327922,lowrey2018plan}.  Our method is also akin to~\cite{cronrathbagger,menda2017dropoutdagger,lee2018safe} where Bayesian extensions of IL are presented to quantify the learner's predictive uncertainty and better guide the queries to the teacher policy. Other Bayesian approaches in policy learning include~\cite{deisenroth2011pilco,gal2016improving,polymenakos2019safe}. ``Frequentist'' alternatives to Bayesian uncertainty quantification also exist~\cite{vovk2005algorithmic,wang2018learning,papernot2018deep,bortolussi2019neural,park2019pac}.

Our work builds on the PLATO IL algorithm~\cite{kahn2017plato} and extends it in three main direction: 1) we consider recurrent architectures, which are more suitable than feedforward ones (used in PLATO) to represent nonlinear state estimation and control; 2) PLATO also derives stochastic policies but, unlike our work, no uncertainty-aware decision-making strategies are considered; 3) PLATO policies do not support systems with external disturbances beyond noise. In our policies instead, random meal disturbances are central.

\section{Conclusion} \label{sec:conclusion}

We introduced a method based on MPC-guided Imitation Learning and Bayesian inference to derive stochastic Neural Network policies for insulin control in an artificial pancreas. Our policies are end-to-end in that they directly map CGM values into insulin control actions. By using Bayesian neural networks, we can crucially quantify prediction uncertainty, information that we incorporate in the insulin therapy decisions. We empirically demonstrated that our stochastic insulin policies outperform traditional MPC with explicit state estimation and are more robust than their deterministic counterparts, as they generalize well to unseen T1D patient parameters and meal disturbance distributions.

\bibliography{mybibfile,pancreas}
\bibliographystyle{icml2019}

\newpage
\appendix
\section*{\Large Appendix}

\vspace{0.3in}
\section{Significance Tests Details}\label{app:sgtest}

Detailed statistics for all control policies and all configurations are available in Table~\ref{tab:full-trajectpry}. We show extra performance measures in this table: $BG_{max}$ and $BG_{min}$, average of maximum and minimum of BG values  (in mg/dL); $u$,  average  insulin  infusion  rate  (in  mU/min).
We performed significance tests out of the collected data to validate our hypotheses regarding the performance of the learned stochastic policies. Details are given below.


\begin{table*}[tbp]
\centering
\caption{Full statistics of the experiments in which virtual T1D patient(s) are regulated by the different controllers/control policies}
\label{tab:full-trajectpry}
\resizebox{\textwidth}{!}{%
\begin{tabular}{@{}clllllll@{}}
\toprule
\multicolumn{1}{l}{}    & Configuration   & $t_{hypo}$ (\%) & $t_{eu}$ (\%) & $t_{hyper}$ (\%) & $BG_{max}$  (mg/dL) & $BG_{min }$(mg/dL) & u (mU/min) \\ \midrule
\multirow{3}{*}{MPC+SI} & Fixed Patient   & 0.00  + 0.00    & 99.84+ 0.60   & 0.16+ 0.60       & 181.28+13.64        & 123.40+3.85        & 28.11+0.95 \\ \cmidrule(l){2-8} 
                        & Varying Patient & 0.00+0.00       & 99.80+0.83    & 0.20+0.83        & 181.55+12.72        & 123.40+4.68        & 28.17+1.05 \\ \cmidrule(l){2-8} 
                        & Patient Cohort  & 0.00+0.00       & 99.15+1.81    & 0.85+1.81        & 184.56+18.06        & 115.82+9.68        & 20.74+4.62 \\ \midrule
\multirow{3}{*}{MPC+SE} & Fixed Patient   & 0.92+2.76       & 80.40+5.33    & 18.68+4.76       & 242.68+12.84        & 84.87+14.68        & 27.98+1.17 \\ \cmidrule(l){2-8} 
                        & Varying Patient & 1.44+3.18       & 79.88+5.46    & 18.69+4.95       & 239.79+12.22        & 81.44+14.80        & 28.16+1.05 \\ \cmidrule(l){2-8} 
                        & Patient Cohort  & 0.85+3.57       & 77.17+15.39   & 21.98+14.30      & 241.17+32.43        & 95.45+21.73        & 20.55+6.52 \\ \midrule
\multirow{3}{*}{DLP}    & Fixed Patient   & 0.32+1.19       & 85.45+4.34    & 14.23+4.00       & 232.63+10.49        & 80.98+9.22         & 27.79+0.78 \\ \cmidrule(l){2-8} 
                        & Varying Patient & 0.53+1.87       & 82.66+6.42    & 16.82+6.25       & 225.66+10.40        & 82.27+10.91        & 28.59+0.76 \\ \cmidrule(l){2-8} 
                        & Patient Cohort  & 0.45+1.91       & 82.63+6.28    & 16.93+6.15       & 225.58+10.41        & 83.14+10.95        & 28.59+0.79 \\ \midrule
\multirow{3}{*}{SLP-M}  & Fixed Patient   & 0.28+0.99       & 86.33+4.17    & 13.44+4.09       & 230.41+10.27        & 81.20+9.31         & 28.39+0.75 \\ \cmidrule(l){2-8} 
                        & Varying Patient & 0.33+1.09       & 86.34+4.20    & 13.33+4.06       & 230.16+10.37        & 81.01+9.77         & 28.38+0.73 \\ \cmidrule(l){2-8} 
                        & Patient Cohort  & 0.21+0.99       & 85.40+4.15    & 14.38+3.90       & 232.72+10.27        & 81.13+8.98         & 28.39+0.72 \\ \midrule
\multirow{3}{*}{SLP-A}  & Fixed Patient   & 0.13+0.64       & 91.73+3.45    & 8.14+3.39        & 216.65+8.05         & 82.42+7.86         & 29.01+0.67 \\ \cmidrule(l){2-8} 
                        & Varying Patient & 0.05+0.27       & 91.73+3.18    & 8.23+3.15        & 216.98+7.68         & 82.50+8.35         & 28.97+0.69 \\ \cmidrule(l){2-8} 
                        & Patient Cohort  & 0.04+0.41       & 85.57+4.12    & 14.39+4.08       & 232.02+8.92         & 86.84+9.58         & 28.61+0.73 \\ \bottomrule
\end{tabular}%
}
\end{table*}


\subsection{ Hypotheses Formulation }
We define our null and alternative hypotheses in the following for each of the performance measures: $t_{hypo}$, $t_{eu}$, $t_{hyper}$, $BG_{max}$, $BG_{min}$ and $u$.

\begin{itemize}
    \item Null hypothesis $ H_0: \mu_{x -y} =0$. The median of the distribution of the difference of a specific performance metric from two sets of experiments regulated by two control policies is zero. Note the hypothesis of zero median of $x-y$ is not equivalent to a hypothesis of equal median of $x$ and $y$. 
    \item Alternative hypothesis $H_A :\mu_{x-y} > 0$. The median of distribution of the differences of a specific performance metric from two sets of experiments regulated by two control policies is larger (smaller) than zero.
\end{itemize}
This formulation is for a paired significance test, which is suitable when comparing, like we do, experimental results about same configuration (and hence, same sampled parameters and disturbances) but different policies.

\subsection{Calculating statistic tests}
For each combination of the compared policies, we use sign tests to test the hypothesis that medians of distribution of the difference between their $t_{hypo}$, $t_{hyper}$ and $BG_{max}$ are larger than zero, and medians of distribution of the difference between their $t_{eu}$, $BG_{min}$ and $u$ are smaller than zero. In the main paper, we set the significance level to $\alpha = 0.005$ (i.e., $99.5\%$ confidence). 
\begin{table*}[htbp]
\centering
\caption{P values of the significance tests for all combinations of compared policies. In each column, the alternative hypothesis is the medians of the observations is larger than zero of the difference in the metrics (i.e., observations from the first policy minus those from the second) in the categories 1) $t_{hypo}$, 2)$t_{hyper}$ and 3) $BG_{max}$, and is smaller than zero in the rest. }
\label{tab:pValues}
\resizebox{\textwidth}{!}{%
\begin{tabular}{@{}clllllll@{}}
\toprule
\multicolumn{1}{l}{Configuration}                   &             & MPC+SE vs SLP-A & DLP vs SLP-A & SLP-M vs SLP-A & MPC+SE vs DLP & MPC+SE vs SLP-M & DLP vs SLP-M \\ \midrule
\multirow{6}{*}{Fixed Patient}                      & $t_{hypo}$  & 5.9090e-03      & 5.0000e-01   & 5.0000e-01   & 2.0695e-02    & 1.3302e-02      & 5.0000e-01   \\ \cmidrule(l){2-8} 
                                                    & $t_{eu}$    & 3.7664e-20      & 7.1194e-20   & 3.8869e-14   & 6.2463e-07    & 2.1929e-10      & 4.2922e-11   \\ \cmidrule(l){2-8} 
                                                    & $t_{hyper}$ & 3.7664e-20      & 1.3449e-19   & 2.6012e-13   & 1.8975e-06    & 9.7412e-10      & 1.3568e-12   \\ \cmidrule(l){2-8} 
                                                    & $BG_{max}$  & 9.8208e-23      & 6.5759e-18   & 6.3970e-16   & 2.0826e-03    & 5.4510e-06      & 5.6818e-08   \\ \cmidrule(l){2-8} 
                                                    & $BG_{min}$  & 8.2858e-01      & 5.4194e-01   & 7.0079e-01   & 9.1488e-01    & 9.7770e-01      & 2.9921e-01   \\ \cmidrule(l){2-8} 
                                                    & $u$         & 1.0000e+00      & 1.0000e+00   & 1.0000e+00   & 9.9897e-01    & 9.9897e-01      & 6.2398e-01   \\ \midrule
\multirow{6}{*}{Varying Patient}                    & $t_{hypo}$  & 2.4617e-05      & 5.8594e-03   & 4.6143e-02   & 1.4725e-02    & 6.2705e-03      & 5.8810e-01   \\ \cmidrule(l){2-8} 
                                                    & $t_{eu}$    & 2.1623e-21      & 6.3970e-16   & 5.4061e-19   & 5.6672e-02    & 3.6187e-10      & 3.8285e-05   \\ \cmidrule(l){2-8} 
                                                    & $t_{hyper}$ & 5.4061e-19      & 5.2608e-15   & 5.4061e-19   & 8.5121e-02    & 2.9933e-07      & 9.3878e-05   \\ \cmidrule(l){2-8} 
                                                    & $BG_{max}$  & 3.7664e-20      & 4.0354e-09   & 3.8869e-14   & 9.7412e-10    & 6.2463e-07      & 9.9598e-01   \\ \cmidrule(l){2-8} 
                                                    & $BG_{min}$  & 5.4194e-01      & 5.6672e-02   & 1.7142e-01   & 3.7602e-01    & 6.2398e-01      & 5.4194e-01   \\ \cmidrule(l){2-8} 
                                                    & $u$         & 1.0000e+00      & 9.9991e-01   & 1.0000e+00   & 9.9257e-01    & 9.9598e-01      & 2.2299e-02   \\ \midrule
\multicolumn{1}{l}{\multirow{6}{*}{Patient Cohort}} & $t_{hypo}$  & 1.9531e-02      & 3.5156e-02   & 1.0742e-02   & 2.9053e-01    & 6.8547e-01      & 7.7275e-01   \\ \cmidrule(l){2-8} 
\multicolumn{1}{l}{}                                & $t_{eu}$    & 1.0301e-03      & 6.6763e-04   & 4.5806e-01   & 1.2305e-01    & 2.1896e-04      & 4.0230e-03   \\ \cmidrule(l){2-8} 
\multicolumn{1}{l}{}                                & $t_{hyper}$ & 1.0301e-03      & 2.7725e-03   & 6.2398e-01   & 1.2305e-01    & 2.1896e-04      & 1.0301e-03   \\ \cmidrule(l){2-8} 
\multicolumn{1}{l}{}                                & $BG_{max}$  & 1.2305e-01      & 9.9792e-01   & 3.7602e-01   & 1.3151e-02    & 3.7602e-01      & 9.9999e-01   \\ \cmidrule(l){2-8} 
\multicolumn{1}{l}{}                                & $BG_{min}$  & 9.9999e-01      & 7.4334e-03   & 2.0826e-03   & 9.9996e-01    & 9.9999e-01      & 7.0079e-01   \\ \cmidrule(l){2-8} 
\multicolumn{1}{l}{}                                & $u$         & 1.0000e+00      & 7.6960e-01   & 9.1488e-01   & 1.0000e+00    & 1.0000e+00      & 4.0230e-03   \\ \bottomrule
\end{tabular}%
}
\end{table*}

\subsection{Conclusion about $H_0$}
We can see the tests reject $H_0$ when comparing $t_{eu}$ and $t_{hyper}$ between SLP-A and all other policies, indicating they are statistical significant larger (smaller) when a fixed or varying patient is regulated by SLP-A than those by other policies. The p-values of the sign test on $BG_{max}$ also support that SLP-A can significantly reduce hyperglycemia for a fixed or varying patient compared to all others. 

In a patient cohort, the null hypothesis cannot be rejected for metrics $t_{eu}$ and $t_{hyper}$ when comparing SLP-A VS SLP-M, indicating there is no significant difference in using these two stochastic policies. However, when comparing these two policies with any other policies, the null is rejected for both $t_{eu}$ and $t_{hyper}$. We can conclude that SLP-A and SLP-M significantly reduce hyperglycemia while maintaining short time in hypoglycemia. 

\section{Comparison Between Supervised Learning and Imitation Learning}\label{app:SL}

\subsection{ Performance improvement}

\begin{figure*}
\centering
\subfloat[Supervised Learning]{\includegraphics[width=.45\linewidth]{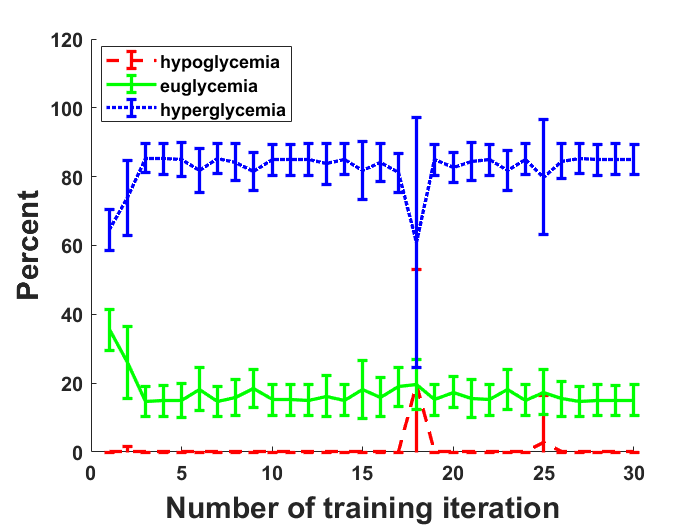}} \hfill
\subfloat[Imitation Learning]{\includegraphics[width=.45\linewidth]{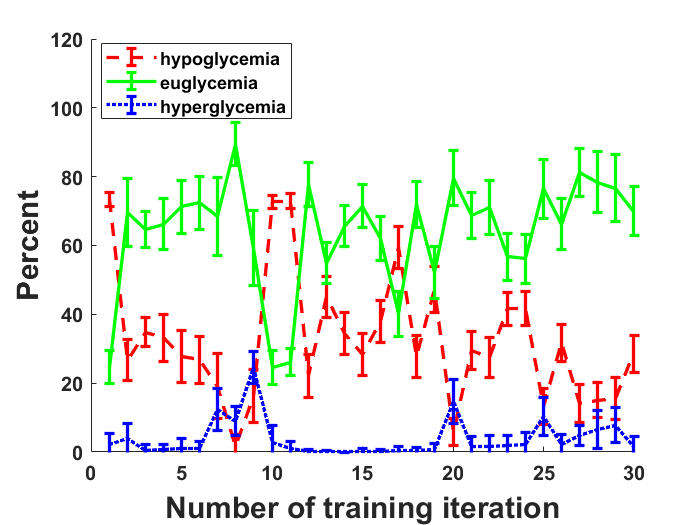}} \hfill
\caption{Mean BG $\pm$ standard deviation on hypoglycemia, euglycemia, hyperglycemia for SLP-A trained by supervised learning v.s. imitation learning. Imitation learning methods explore the state space whose distribution that has a small distance to that in test time, so the SLP-A improves faster with IL-based methods.}
\label{fig:ILvsSL}
\end{figure*}

 Figure~\ref{fig:ILvsSL} (a) show the proportion of time blood glucose is in every state (eu-, hyper- and hypo-glycemia) at each training iterations using supervised learning, which improves much slower than Figure~\ref{fig:ILvsSL} (b).

We can see the learner policy trained by IL outperforms it by SL in reducing hypo-/hyper-glycemia and keeping patients' blood glucose in healthy range (euglycemia). The mean of maximum blood glucose decreases faster while the mean of minimum blood glucose increases to prevent dangerous hypoglycemia when SLP-A is trained by IL methods.

\begin{figure}[tbp]
\centering
\includegraphics[width=0.85\columnwidth]{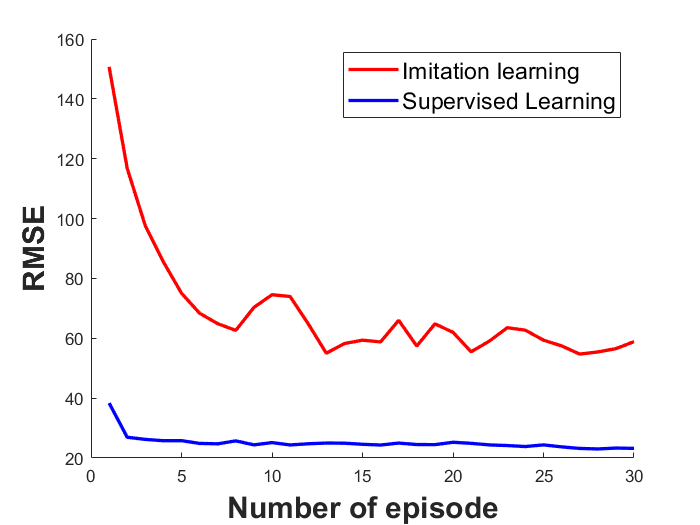}
\caption{RSME of SL and IL at each training iteration.}
\vspace{-.3cm}
\label{fig:RSME}
\end{figure}

Figure~\ref{fig:RSME} is the root mean squared error (RMSE) of both learning methods at each training iteration. As we can see from the curves, the training RMSE for SL and IL goes down as the number of training episodes increases. The RMSE of IL starts larger but decrease dramatically especially after the first few episodes.
Although the RMSE is smaller, stochastic learner policy trained by SL methods has a limited training state space and the out-of-distribution disturbance leads to slow learning process from supervision (i.e., MPC controller with perfect state information).

\section{Uncertainty Analysis of Stochastic Learner Policy}
\label{app:uncertain}
In our machine learning model for the stochastic learner policy, we quantify the uncertainty of our predictive action. Quantifying the uncertainty is critical in real-world settings such as artificial pancreas (AP) system, in which often involve out-of-distribution disturbance and patient variations. We analyze the uncertainty quantifying ability of our Monte-Carlo Dropout Bayesian method to investigate the effect of covariate shift and virtual patient model variation. 

\subsection{Coefficients of variation}

The coefficients of variation (CV), aslo known as relative standard deviation, is defined as the division of the standard deviation and the mean. 

\begin{figure}[tbp]
\centering
\includegraphics[width=0.85\columnwidth]{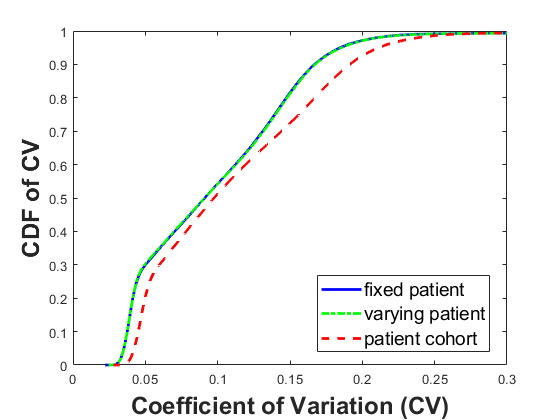}
\caption{CDFs of Coefficients of variations of the stochastic learner policy outputs under three different patient configuration. The uncertainty of the patient model caused by intra-/inter-patient parameters variations is the largest in a patient cohort and smallest in a fixed virtual patient.}
\label{fig:cov}
\end{figure}

Figure~\ref{fig:cov} is the cumulative distribution functions (CDF) of the CV of the stochastic learner policy outputs when the learner policy is the control algorithm of an AP system under three patient configuration as illustrated in Section~6 of our paper. The curve of patient cohort experiments is the slowest to reach a probability of 1, and the curve of the fixed patient experiments is the fastest but the curve of varying patient experiment is close. We remark that, in this case, a faster growing CDF implies a higher probability of smaller CV values. 


\subsection{Kolmogorov-Smirnov test}

In statistics, the Kolmogrov-Smirnov test, also known as K-S test, tests null hypothesis for the equality of two probability distributions, or a sample distribution with a reference probability distribution. 

We use K-S test to quantify the distance between the empirical CDF of the CV of SLP-A outputs under three patient configurations with significance level $\alpha = 0.05$. The null hypotheses are rejected when compare the empirical CDF of CV from patient cohort experiment and others in 5\% confidence level. Although the CDf curve of fixed patient experiment is faster but the K-S test between it and the CDF of varying patient experiment returns a decision that the null hypothesis is accepted with p value 0.2453, thus they are likely from the same CDF. The p values are believed to be accurate since the sample sizes of the CDFs are large enough, such that $(n_1 *n_2)/(n_1+n_2) \gg 4 $, with $n_1$ and $n_2$ are the sample sizes of compared CDFs.

\section{Generalization to different diet behaviors}\label{app:meal}

The stochastic learner policy can not only generalize its control ability onto different virtual patients but also to different meal distributions. 

\subsection{Meal distribution}

The attributes of the distinct meal distribution we used to evaluate the generalization ability of our stochastic learner policy is summarized in Table~\ref{table:meals}. The new distribution has a higher probability of occurrence of snacks and larger snacks with larger carbohydrate amounts. This describes a larger system disturbance to the virtual patient models, and is more challenging for our stochastic learner policy to regulate the patients' blood glucose level.   

\begin{table}[htbp]
\centering
\caption{Attributes of a different meal disturbance distribution during testing to that during training. CHO amounts and starting times are sampled uniformly from the reported intervals.}
\label{table:meals}
\resizebox{\linewidth}{!}{%
\begin{tabular}{lllllll}
\hline
                 & breakfast              & snack 1                & lunch                   & snack 2                & dinner                 & snack 3                \\ \hline
Occurrence       & \multirow{2}{*}{100}   & \multirow{2}{*}{80}    & \multirow{2}{*}{100}    & \multirow{2}{*}{80}    & \multirow{2}{*}{100}   & \multirow{2}{*}{80}    \\
Probability (\%) &                        &                        &                         &                        &                        &                        \\ \hline
CHO Amount       & \multirow{2}{*}{40-60} & \multirow{2}{*}{15-30} & \multirow{2}{*}{70-110} & \multirow{2}{*}{15-30} & \multirow{2}{*}{55-75} & \multirow{2}{*}{15-30} \\
(gram)           &                        &                        &                         &                        &                        &                        \\ \hline
Time of          & 3:00-                  & 7:00-                  & 10:00                   & 14:00                  & 17:00                  & 21:00                  \\
day (hour)       & 7:00                   & 10:00                  & 14:00                   & 17:00                  & 21:00                  & 23:00                  \\ \hline
\end{tabular}%
}
\end{table}

\subsection{Performance}

We also test our stochastic learner policy in three patient configurations when the patients' meals are subject to the distribution in Table~\ref{table:meals}. The results of the performance are shown in Table~\ref{table:meals_results}.

\begin{table}[htbp]
\centering
\caption{The full performance of SLP-A with unseen disturbances.}
\label{table:meals_results}
\resizebox{\linewidth}{!}{%
\begin{tabular}{@{}llll@{}}
\toprule
\multicolumn{1}{c}{\multirow{2}{*}{\begin{tabular}[c]{@{}c@{}}Performance\\ Measures\end{tabular}}} & \multirow{2}{*}{\begin{tabular}[c]{@{}l@{}}Fixed Patient\\ Configuration\end{tabular}} & \multirow{2}{*}{\begin{tabular}[c]{@{}l@{}}Varying Patient\\ Configuration\end{tabular}} & \multirow{2}{*}{\begin{tabular}[c]{@{}l@{}}Patient Cohort\\ Configuration\end{tabular}} \\
\multicolumn{1}{c}{}                                                                               &                                                                                        &                                                                                          &                                                                                         \\ \midrule
$t_{hypo}$ (\%)                                                                                    & 0.35$\pm$1.08                                                                              & 0.67$\pm$1.83                                                                                & 0.00$\pm$0.00                                                                               \\ \midrule
$t_{eu}$ (\%)                                                                                      & 91.41$\pm$3.67                                                                             & 90.77$\pm$4.44                                                                               & 85.92$\pm$3.96                                                                              \\ \midrule
$t_{hyper}$ (\%)                                                                                   & 8.24$\pm$3.43                                                                              & 8.56$\pm$3.73                                                                                & 14.08$\pm$3.96                                                                              \\ \midrule
$BG_{max}$ (mg/dL)                                                                                 & 218.25$\pm$8.05                                                                            & 217.03$\pm$6.70                                                                              & 233.22$\pm$10.48                                                                            \\ \midrule
$BG_{min }$(mg/dL)                                                                                 & 76.52$\pm$6.02                                                                             & 76.15+6.70                                                                               & 87.69$\pm$9.30                                                                              \\ \midrule
u (mU/min)                                                                                         & 29.26$\pm$0.75                                                                             & 29.32$\pm$0.74                                                                               & 28.70$\pm$0.67                                                                              \\ \bottomrule
\end{tabular}%
}
\end{table}

\subsection{Significance tests}

We use above described sign test to evaluate, for each configuration, if there are significant performance differences when training-time VS never-before-seen disturbances are used. 
The two sets of observations are from Table~\ref{table:meals_results} and those from SLP-A in Table~\ref{tab:full-trajectpry}. 
The p values of the statistical tests are listed in Table~\ref{table:pvalue_meal}. 

\begin{table}[htbp]
\centering
\caption{P values of sign tests for experiments with SLP-A under trained meal distribution VS unseen meal distribution.}
\label{table:pvalue_meal}
\resizebox{\linewidth}{!}{%
\begin{tabular}{@{}llll@{}}
\toprule
\multicolumn{1}{c}{\multirow{2}{*}{\begin{tabular}[c]{@{}c@{}}Performance\\ Metrics\end{tabular}}} & \multirow{2}{*}{\begin{tabular}[c]{@{}l@{}}Fixed Patient\\ Configuration\end{tabular}} & \multirow{2}{*}{\begin{tabular}[c]{@{}l@{}}Varying Patient\\ Configuration\end{tabular}} & \multirow{2}{*}{\begin{tabular}[c]{@{}l@{}}Patient Cohort\\ Configuration\end{tabular}} \\
\multicolumn{1}{c}{}                                                                               &                                                                                        &                                                                                          &                                                                                         \\ \midrule
$t_{hypo}$ (\%)                                                                                    & 5.9235e-02                                                                             & 2.2125e-03                                                                               & 1.0000e+00                                                                              \\ \midrule
$t_{eu}$ (\%)                                                                                      & 2.9921e-01                                                                             & 3.6275e-02                                                                               & 7.6960e-01                                                                              \\ \midrule
$t_{hyper}$ (\%)                                                                                   & 4.5806e-01                                                                             & 2.3040e-01                                                                               & 7.0079e-01                                                                              \\ \midrule
$BG_{max}$ (mg/dL)                                                                                 & 3.7602e-01                                                                             & 5.4194e-01                                                                               & 4.5806e-01                                                                              \\ \midrule
$BG_{min }$(mg/dL)                                                                                 & 5.6818e-08                                                                             & 1.4832e-05                                                                               & 8.2858e-01                                                                              \\ \midrule
u (mU/min)                                                                                         & 2.2299e-02                                                                             & 2.2299e-02                                                                               & 2.9921e-01                                                                              \\ \bottomrule
\end{tabular}%
}
\end{table}


We can observe that all p-values in the patient cohort experiments are very high (the smallest is $0.29921$), indicating that, as expected, no significant performance difference is observed in patient cohort: SLP-A is able to generalize its control ability to an unseen meal distribution during training phase. In the other two patient configurations, the null hypotheses are also accepted except $t_{hypo}$ in the second column and both $BG_{min}$.

\vspace{12pt}

\end{document}